\documentclass[10pt,twocolumn]{article}

\usepackage[a4paper,left=0.85in,right=0.85in,top=0.9in,bottom=0.9in,columnsep=0.3in]{geometry}
\usepackage{amsmath,amssymb,amsfonts}
\usepackage{algorithm}
\usepackage{algpseudocode}
\usepackage{graphicx}
\usepackage{textcomp}
\usepackage{booktabs}
\usepackage{url}
\usepackage[hidelinks]{hyperref}

\begin{document}

\twocolumn[
{
\centering
{\LARGE \bfseries ENHANCED TRADING SYSTEMS WITH INTELLIGENT MARKET ANALYSIS \par}
\vspace{10pt}
{\large
1 Vedanth Padmaraman, 1 Shereen Anand, 1 Vamshi Vishruth J, 1 Vineet V Pai,\\
2 Arvind Srinath K.T, 2 Dr. Subramanyam Natarajan, 1 Dr. Shylaja S S
\par}
\vspace{8pt}
{\small
1 Department of Computer Science and Engineering, PES University, Bengaluru, India\\
2 Key Resource Person, Bengaluru, India
\par}
\vspace{6pt}
{\small
Email: vedanthpadmaraman0566@gmail.com, shreenanand@gmail.com, vamshivishruthj@gmail.com, ho75vineet@gmail.com,\\
shylajass@pesu.edu, arvind.srinath@gmail.com, natarajan@pes.edu
\par}
\vspace{10pt}
\hrule
\vspace{8pt}
\begin{minipage}{\textwidth}
\noindent\textbf{Abstract:} This paper proposes a hybrid reinforcement learning trading framework integrating intraday market data with transformer-based financial sentiment signals. The system extracts sentiment dynamics and semantic event clusters from financial news and incorporates them into a PPO-based trading agent. Experiments on intraday NSE data demonstrate improved risk-adjusted returns compared to price-only baselines.

\vspace{6pt}
\end{minipage}
\vspace{8pt}
\hrule
\vspace{14pt}
}
]

\section{Introduction}

The present-day financial markets are understood not only by numerical signals such as price and volume but also by textual narratives such as news, analyst notes, social networks, etc. Standard algorithmic trading approaches handle these signals separately. This study bridges them by (1) use of the FinBERT pipeline with fallback lexicon scoring to extract textual sentiment, (2) deriving higher-order sentiment dynamics (velocity, acceleration, correlation), and (3) training RL agents across a controlled curriculum to exploit these signals together with numerical signals.

This work develops a hybrid trading framework that uses the dynamics associated with textual data and news embedding-based structure in the decision-making process of reinforcement learning (RL) agents. Our objective was to establish the methodological foundations required to (i) extract temporally aligned sentiment features, (ii) quantify sentiment evolution through physically inspired derivatives, and (iii) evaluate whether such features contain predictive structure that can be exploited by RL agents.

Our contributions:
\begin{enumerate}
    \item A module that synchronized OHLCV data with asynchronous signals from news data, including clustering/de-duplication.
    \item Formulated MDP designs for multi-RL environment tiers, including observation/action spaces and reward engineering that explicitly integrates sentiment.
    \item A practical PPO training procedure with training, checkpoint, and logs.
    \item Detailed experimental plan, metrics tests, and reproducibility checklist.
\end{enumerate}

Here, we introduce two categories of textual signals along with the numerical indicators.

\textbf{Sentiment velocity:} In addition to raw sentiment scores, sentiment velocity (the first-order temporal derivative of sentiment, representing the rate of sentiment change) and sentiment acceleration (the derivative of sentiment velocity) were calculated. These higher-order features aim to capture the momentum of market narratives, and therefore assess the rapid shifts that could occur in market sentiment by judging the textual tone, and hence anticipate subsequent actions to be taken by the self-trading agent to achieve higher returns.

\textbf{Embedding-driven news structure:} Dense semantic embeddings for each news article were constructed, and exploratory clustering was performed to identify groups of thematically related events. Preliminary analysis using future return windows of different timeframes (1 hour, 4 hours, 1 day, 3 days) revealed that certain clusters exhibit consistent directional impacts on returns. However, substantial event duplication bias was also identified, where clusters were dominated by near-identical articles surrounding a single financial/corporate event, artificially inflating the predictive signals backed by the clusters.

To mitigate this, we followed a rigorous de-biasing pipeline:
\begin{enumerate}
    \item De-duplicated articles into canonical events.
    \item Re-clustered events using HDBSCAN and K-Means with cluster-size constraints.
    \item Transitioned from sign-based return analysis to quantile-binned returns with bootstrap confidence intervals.
    \item Computed permutation-based significance tests and Spearman information coefficients.
    \item Generated event-study and horizon-wise heatmaps to evaluate the robustness of cluster signals. This process isolated only those clusters whose return patterns generalized beyond single-event artifacts.
\end{enumerate}

The validated textual features, sentiment derivatives, statistically reliable cluster indicators, and news-event embeddings were then integrated into three progressively expressive RL environments, which were designed to test whether policy performance improves when agents have access to higher-order market narratives. We formalized Markov Decision Processes (MDPs), developed sentiment-aware reward shaping functions, and trained agents using Proximal Policy Optimization (PPO) with a portfolio-state-aware policy wrapper. Experiments on intraday datasets (e.g., BHEL, Jan--Jul 2020) demonstrated that incorporating these textual signals yielded measurable gains in return, risk-adjusted metrics, and stability under curriculum training.

\section{Related Work}

This work was synthesized from incremental work along three related research strands: (i) reinforcement-learning (RL) methods for algorithmic and intraday trading, (ii) financial sentiment analysis including transformer-based models and lexicon methods, and (iii) multimodal/hybrid decision approaches that aimed at combining numeric market data with textual signals. We contextualize the representative previous work, highlight common limitations, and identify the gaps our paper addresses along each of the strands.

\subsection{Reinforcement Learning for Trading}

The research thus far along this path treated trading as a sequential decision problem and utilized RL to learn trading policies directly from market data. Frameworks such as \textit{FinRL} provide modular environments, state/action abstractions, and standard algorithms for single-asset and portfolio problems [5]. Recent algorithmic advances emphasize instability-aware training, risk-sensitive objectives, and architectures suited to intraday tasks: for example, \textit{DeepScalper} proposed an architecture tailored to fleeting intraday opportunities, combining fine-grained action branching, temporal encoders, and auxiliary risk targets to balance profit and volatility [12]. More recently, mixture-of-actor approaches employed multiple specialized policies (actors) and optimal transport or other regularizers to differentiate market regimes and allocate learning resources more effectively [10]. These RL works demonstrate that carefully designed environments, reward formulation, and architectures affect intraday performance, and the agent can adapt to different such patterns.

\textbf{Limitations.} Despite their strengths, the majority of RL-for-trading research relies predominantly on numeric inputs (prices, volumes, order-book features, technical indicators). This numeric focus makes it difficult for such agents to anticipate external information shocks originating from textual sources (news, disclosures, social commentary) that can materially affect short-term price dynamics.

\subsection{Financial Sentiment Analysis and LLMs}

Financial sentiment modeling has advanced from lexicon-based scoring to transformer-based models, either pre-trained or fine-tuned on finance-related data. Models such as FinBERT and similar domain-tuned transformers produce more accurate polarity and intensity estimates for financial text than generic models or lexicons [14]. More recent work explores aligning large language models (LLMs) with human instructions and market feedback, combining instruction tuning, retrieval augmentation, and reinforcement or feedback loops to improve sentiment signals that better reflect market reactions [11]. Empirical studies showed that LLM-derived sentiment can correlate with subsequent returns and be useful for supervised prediction or rule-based strategies.

\textbf{Limitations.} Prior sentiment-based work typically stops short of integrating these signals with other principled RL decision frameworks. Moreover, many studies treated sentiment as a static score at an article timestamp, rarely exploiting higher-order temporal derivatives (e.g., velocity/acceleration) or addressing clustering and event-duplication biases that can inflate apparent predictive power. The asynchronous nature of news relative to high-frequency OHLCV data and the problem of near-duplicate reporting were often, but not systematically, addressed in downstream decision experiments.

\subsection{Multimodal and Hybrid Decision Systems}

There is growing interest in combining multiple modalities (numeric and textual) for financial prediction. Several supervised and hybrid methods fuse sentiment features with price/technical indicators for return forecasting or risk estimation. A smaller set of works attempts to incorporate textual signals into trading strategies or RL environments, demonstrating improved performance in some settings. However, many such attempts were exploratory: they do not (i) provide rigorous statistical de-biasing of news clusters, (ii) formalize MDPs that explicitly accommodate asynchronous textual events, or (iii) evaluate the contribution of structured textual derivatives (velocity, acceleration) and embedding-cluster signals under robust inference (bootstrap CIs, permutation tests, information coefficients).

\textbf{Limitations and open gaps.} Existing hybrid approaches rarely combine the following elements in one methodological pipeline: de-duplication of events, embedding-based event clustering with cluster-level inference, higher-order sentiment derivatives, and a reproducible RL experimentation protocol (MDP formalization, curriculum training, ablation studies). Consequently, there exists a lack of clarity on when and how text-derived features truly add sequential decision value beyond price-only baselines.

\subsection{Positioning and Contributions}

Against this backdrop, our paper makes a methodological proof-of-concept contribution rather than presenting a deployment system. Concretely, we (i) compute and validate \textit{sentiment derivatives} (velocity and acceleration) as candidate predictive features across multiple forward horizons, (ii) construct a de-biased event representation pipeline (near-duplicate de-duplication, embedding clustering, cluster robustness testing), and (iii) integrate the validated textual features into formally defined RL environments with sentiment-aware observation and reward structures. By combining rigorous statistical validation of textual signals with principled MDP and curriculum design, our approach aims to fill the gap between sentiment-centered predictive work and RL-based trading research. We therefore position our contribution as an incremental but nontrivial extension of the RL trading literature --- one that introduces multimodal, statistically validated textual features into the RL decision loop.

\section{Dataset and Preprocessing}

This section describes the construction of the multimodal dataset used for training and evaluating our hybrid trading agents. We integrated high-frequency OHLCV market data with timestamped financial news by applying a unified preprocessing pipeline to ensure temporal consistency, reproducibility, and statistical reliability. All signals used by the RL agents---including technical indicators, sentiment scores, sentiment derivatives, and cluster/event features---were calculated from scratch using our own processing scripts.

\subsection{Market Data}

We collect intraday OHLCV data for two actively traded Indian equities: \textbf{BHEL} and \textbf{JSW Infrastructure}. Data were scraped at a resolution of 1 minute from March 2025 to August 2025. Each per-instrument file contains the canonical schema: timestamp, open, high, low, close, volume.

\textbf{Data integrity checks} include:
\begin{itemize}
    \item Strict chronological ordering and duplicate-timestamp removal,
    \item Normalizing timezone to IST (UTC+5:30),
    \item Removal of non-market-hour records,
    \item Sanity checks on volume spikes, outliers, and malformed candles.
\end{itemize}

\subsection{Train--Test Protocol}

To ensure fair evaluation and reduce temporal leakage, we adopted a rolling-window split:
\begin{itemize}
    \item \textbf{Training window:} First 3 months of each instrument's time range (e.g., March--May 2025).
    \item \textbf{Testing window:} Subsequent 3 months (e.g., June--August 2025).
\end{itemize}

Each RL experiment was run independently per instrument using this 3+3 month split, allowing for cross-instrument generality analysis while preserving chronological causality.

\subsection{Resampling and Normalization}

Despite the raw data being recorded at a frequency of 1 minute, we constructed multiple synchronized resolutions for downstream components:
\begin{itemize}
    \item \textbf{1-minute bars} for high-frequency RL training,
    \item \textbf{5-minute and 1-hour bars} for smoothed evaluation, baselines, and sentiment-aligned sanity checks.
\end{itemize}

For each resampled bar, the OHLC values were aggregated using standard conventions, and a summation of volume was carried out. Missing bars are filled forward for price columns and zero for volume, where appropriate. All numeric features were standardized using \textbf{rolling z-score normalization} computed strictly within the training window to avoid look-ahead bias.

\subsection{Technical Indicator Computation}

We computed a comprehensive set of technical features per bar, including:
\begin{itemize}
    \item \textbf{Trend and smoothing:} SMA/EMA (5, 20, 50),
    \item \textbf{Momentum:} MACD (12, 26, 9), RSI (14),
    \item \textbf{Volatility:} rolling standard deviation (10, 20),
    \item \textbf{Liquidity:} log(volume+1), rolling volume Z-score,
    \item \textbf{Price microstructure:} logarithmic returns, realized volatility, high--low spread, close--open spread.
\end{itemize}

All indicators were computed using our own scripts to ensure reproducibility and maintain a controlled feature-generation pipeline.

\subsection{News and Textual Data}

We scraped timestamped financial news and headlines that matched the selected instruments during the same period (March--August 2025). Each article contains:
\begin{itemize}
    \item timestamp,
    \item headline or full body text,
    \item source metadata,
    \item optional unique IDs.
\end{itemize}

Raw text was minimally cleaned: converting to lowercase, removal of HTML tags, and retention of domain-specific tokens (e.g., EBITDA, MoU, ATF, subsidy).

\subsection{Event De-duplication and News-Clustering Pipeline}

To prevent statistical irregularities arising from repeated rewrites of the same corporate event, we constructed a structured de-duplication and clustering pipeline:
\begin{enumerate}
    \item \textbf{Semantic embedding:} Each article was encoded using an LLM-based sentence embedding model, all-MiniLM-L6-v2.
    \item \textbf{Initial deduplication:} Within sliding 1--4 hour windows, we clustered the embeddings using cosine similarity (threshold $\approx 0.80$--$0.85$) to collapse identical or nearly identical rewrites.
    \item \textbf{Event representation:} For each cluster, we retained the highest-quality representative article and treated it as the canonical `event'.
    \item \textbf{Global event clustering:} Canonical events were further re-clustered over multi-day windows using HDBSCAN to identify higher-level thematic clusters (e.g., earnings, regulatory announcements, management changes).
\end{enumerate}

This two-stage process was used to remove event duplication bias and produces stable event clusters for statistical validation.

\subsection{Sentiment Processing and Derivative Signals}

For each canonical event, the sentiment was computed using a two-stage fallback system:
\begin{itemize}
    \item \textbf{Primary:} FinBERT-based sentiment classifier,
    \item \textbf{Fallback:} Domain-tuned lexicon scoring for low-confidence cases.
\end{itemize}

Further temporal derivatives were computed for the sentiment:
\begin{itemize}
    \item \textbf{Sentiment velocity:} change in sentiment score per unit time,
    \item \textbf{Sentiment acceleration:} change in sentiment velocity.
\end{itemize}

These features were then aligned with the nearest market-time bins and synchronized with the OHLCV data.

\subsection{Multimodal Alignment for RL}

Lastly, all numeric and textual signals were fused into MDP-ready state vectors. Alignment steps include:
\begin{itemize}
    \item timestamp quantization to 1-minute bars,
    \item forward propagation of event-based features until the next event,
    \item masking of unavailable features during training to avoid leakage,
    \item standardization of features per instrument.
\end{itemize}

\section{Reinforcement Learning Environment}

This section describes the formulation of the reinforcement learning (RL) environment with which the trading agent interacts. The RL environment is a crucial component in any RL-based system, as it defines the agent's capabilities, restrictions, and learning objective.

Our environment is modeled around a time series of financial data. At each timestep, the agent consumes an observation and outputs an action. A reward is assigned for each timestep, and the training process aims to optimize the cumulative reward over many timesteps.

\subsection{Observation Space}

The observation space is the set of values that the agent can consume as input. We incorporate, in our observation space, a variety of technical indicators characterizing positional information as well as asset performance. Formally, each observation is a tuple consisting of the following fields:
\begin{itemize}
    \item Technical Indicators (normalized so that maximum magnitude is 1):
    \begin{itemize}
        \item Relative Strength Index (RSI)
        \item Average Directional Index (ADX)
        \item Ultimate Oscillator (ULTOSC)
        \item William's \%R (WILL\%R)
    \end{itemize}
    \item Positional Parameters:
    \begin{itemize}
        \item Current position: whether the agent is holding a short, neutral, or long position, encoded as $-1$, $0$, or $1$
        \item Current capital
        \item Fraction of trading day remaining
    \end{itemize}
\end{itemize}

We also incorporate frame stacking, wherein a fixed number of past observations are stacked together and passed as input to the neural network. This is a simple measure that allows the agent to learn deeper temporal patterns.

The indicators are constructed from raw financial data, as discussed in Section 3.

\subsection{Action Space}

The action space is the set of values that the agent can output. We use an action space consisting of 2 binary signals:
\begin{itemize}
    \item \textbf{Should-Sell:} This indicates that a share should be sold.
    \item \textbf{Should-Buy:} This indicates that a share should be bought.
\end{itemize}

In a long position, Should-Sell = 1 causes a share to be sold. Conversely, in a short position, Should-Buy = 1 causes a share to be bought. In a neutral position, if only one of the signals is 1, that signal is heeded (i.e., Should-Sell causes a sale, Should-Buy causes a buy).

\subsection{Reward Function}

The reward function is derived from the change in portfolio value. At each timestep $t$, the reward function is defined as
\begin{equation*}
    r_t = p_t - p_{t-1}
\end{equation*}
where $p_t$ is the portfolio value, defined by
\begin{equation*}
    p_t = n_t s_t + c_t
\end{equation*}
where $s$ is the share price, $n$ is the number of shares held, and $c$ is the capital.

\begin{figure}[t]
    \centering
    \includegraphics[width=\linewidth]{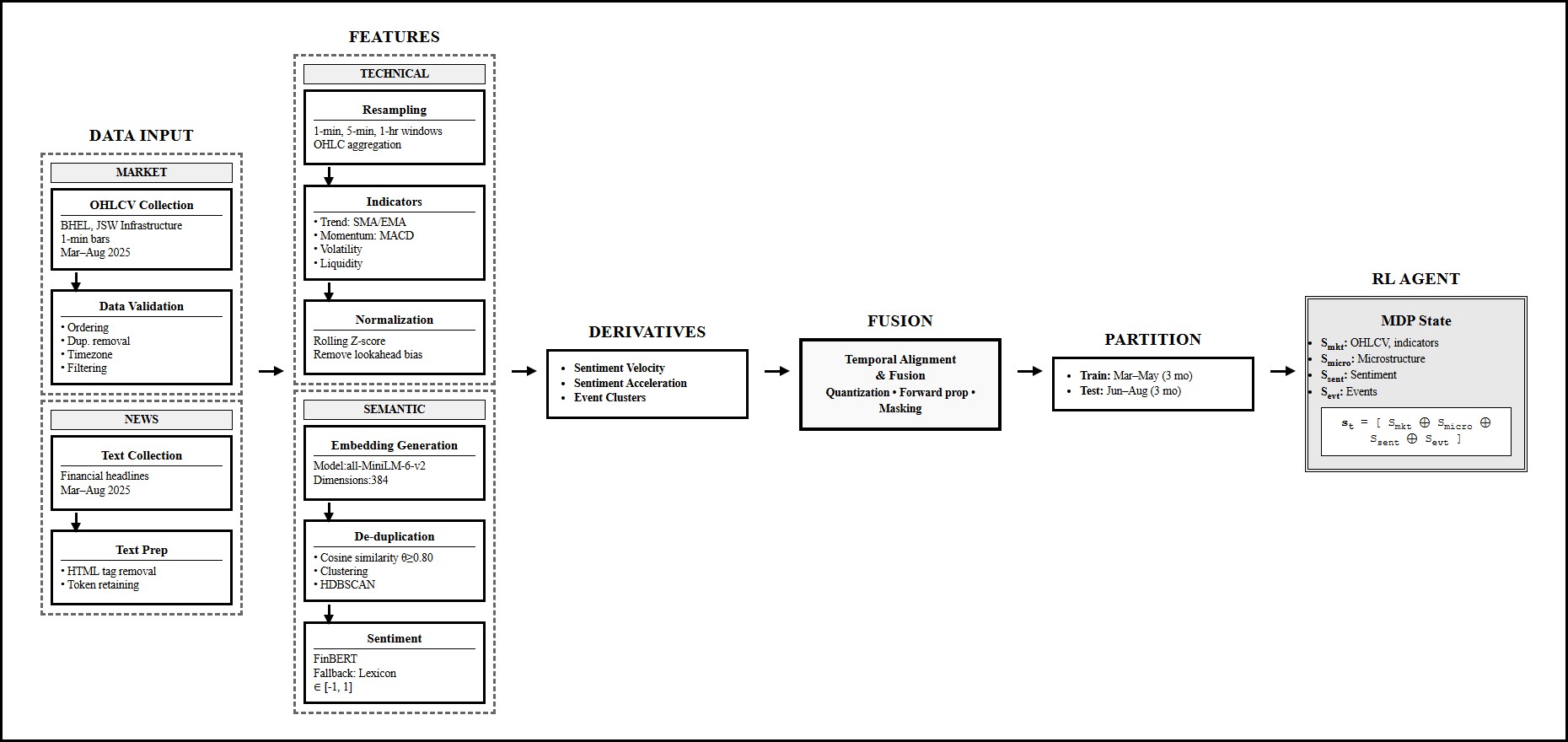}
    \caption{Architecture of the Proposed System}
    \label{fig:architecture}
\end{figure}

\section{Statistical Analysis of Sentiment and Semantic Signals}

\subsection{Sentiment Dynamics and Predictive Strength}

To assess whether textual information contains exploitable structure beyond raw polarity, we evaluated the temporal derivatives of aggregated sentiment. Consistent with our hypothesis, it was observed that sentiment acceleration --- the second-order change in sentiment score --- exhibited a stronger correlation with forward returns than raw sentiment or sentiment velocity.

Across both instruments, acceleration peaks aligned with elevated volatility and directional moves, suggesting that rapid shifts in narrative tone preceded the corresponding price adjustments. Importantly, acceleration produced a statistically significant median-return separation in the 4-hour and 12-hour forward horizons, whereas raw sentiment scores showed weaker or inconsistent effects.

Thus, sentiment acceleration improved responsiveness, capturing the speed at which markets internalize textual information rather than its absolute polarity. This supported its inclusion as a Level-3 feature within the RL state space.

\section{System Architecture}

As shown in Fig.~\ref{fig:architecture}, the proposed model integrates multimodal inputs.

\subsection{Semantic Embeddings Reveal Non-Obvious Structure}

Although sentiments captured the directional tone, they ignored the underlying semantics of the event itself. Using LLM-derived embeddings of canonical news events, we performed clustering to identify semantically coherent groups (e.g., earnings, regulatory actions, management changes, sector-wide macro catalysts).

Our analysis showed that embedding clusters revealed patterns that could not be observed through sentiment alone. Clusters with nearly identical sentiment distributions nevertheless show markedly different future return characteristics, indicating that coarse polarity measures are insufficient to capture nuanced market responses.

This validates the choice of embeddings as a complementary channel---not just noise filtering or de-duplication, but an information dimension that identified recurring thematic contexts that sentiment models under-differentiate.

\subsection{Cluster-Level Return Patterns and Market Regimes}

Across instruments, we identified between 5 and 8 stable event clusters after de-duplication and significance filtering. Using quantile-binned forward returns and bootstrapped 95\% confidence intervals, we observed:
\begin{itemize}
    \item Clusters 3 and 7 consistently exhibited positive median returns throughout the 4--12 hour horizons, with statistically separated confidence bounds.
    \item Event clusters dominated by highly specific one-off events were filtered out via de-duplication and cluster-size thresholds, reducing the event-duplication bias.
    \item Significance tests (Spearman IC, permutation tests) further confirmed that certain clusters carry robust predictive information under resampling.
\end{itemize}

We interpreted these clusters as semantic market regimes, reflecting distinct economic narratives rather than price-driven regimes (as in MOT-style RL). These findings demonstrated that semantic similarity between events is a valuable predictor of how the market will respond in the near future.

\subsection{Optimal Signal Density at 4-Hour Aggregation}

We evaluated returns across multiple horizons (1 hour, 4 hours, 1 day, 3 days). The results showed that the 4-hour bar interval consistently provides the most stable and dense predictive signal for both sentiment acceleration and embedding-cluster features.

Shorter horizons (1 hour) captured noise and transient microstructure effects, while longer horizons (1 day, 3 days) diluted the impact of individual events due to overlapping narratives. The 4-hour horizon balances rapidity with statistical reliability, leading to the clearest separation between cluster-level return distributions.

This informs both our feature-engineering decisions and the granularity of the RL environment's observation.

\subsection{Sentiment-Guided Action Modulation}

One of the key innovations of our approach was the dynamic modulation of raw RL actions based on sentiment signals. The modulated action $\tilde{a}_t$ was calculated as:
\begin{equation}
\tilde{a}_t =
\begin{cases}
\min\big(a_t(1 + |s_t| c_t),\, 1.0\big) & \text{if } \operatorname{sign}(a_t) = \operatorname{sign}(s_t) \\
& \text{and } |s_t| > \theta \\[4pt]
a_t(1 - |s_t| c_t) & \text{if } \operatorname{sign}(a_t) \neq \operatorname{sign}(s_t) \\
& \text{and } |s_t| > \theta \\[4pt]
a_t & \text{otherwise}
\end{cases}
\end{equation}
where $\theta$ is a sentiment threshold (empirically set to 0.2). This modulation implements two important behaviors:
\begin{enumerate}
    \item \textbf{Amplification:} When the agent's intended action aligns with strong sentiment signals, the position size is increased proportionally to sentiment magnitude and confidence.
    \item \textbf{Attenuation:} When the agent acts against prevailing sentiment, the position is reduced to limit exposure to sentiment-driven price movements.
\end{enumerate}

\section{Evaluation Protocol and Metrics}

Our proposed multimodal trading framework was evaluated using standard quantitative finance metrics, RL-specific validation procedures, and statistically robust comparisons across model variants. Because the goal of this work was to demonstrate methodological viability rather than production-grade deployment, the evaluation emphasized risk-adjusted performance, return distribution characteristics, and statistical reliability of the observed improvements.

\subsection{Trading Performance Metrics}

The following performance measures were calculated in the 3-month test window for each instrument:

\begin{itemize}
    \item \textbf{Cumulative Return (ROI):} Measured the total multiplicative growth of the portfolio over the test horizon. This was considered the primary measure of directional profitability.
\end{itemize}

\textbf{Sharpe Ratio:}
\begin{equation*}
    \text{Sharpe} = \frac{E[r]}{\sigma[r]}
\end{equation*}
where $r$ refers to the daily returns. Because test windows are short and risk-free rates are negligible at this horizon, we assumed a risk-free value of zero. This reflects the efficiency of the reward-to-variability ratio.

\textbf{Sortino Ratio:}
\begin{equation*}
    \text{Sortino} = \frac{E[r]}{\sigma[r_{r<0}]}
\end{equation*}
This ratio highlights whether the improvements arise from controlled downside risk rather than symmetric variance.

\begin{itemize}
    \item \textbf{Maximum Drawdown (MDD):} This measure captures the worst decline from peak-to-trough in the portfolio. Lower MDD indicates improved risk management and resilience to loss streaks.
\end{itemize}

\textbf{Win Rate:}
\begin{equation*}
    \text{WinRate} = \frac{\#\text{winning trades}}{\#\text{total trades}}
\end{equation*}

\textbf{Profit Factor:}
\begin{equation*}
    \text{ProfitFactor} = \frac{\text{gross profit}}{\text{gross loss}}
\end{equation*}
These metrics provide interpretable diagnostics on the agent's trade-level behavior.

\begin{itemize}
    \item \textbf{Return Stability:} We report the standard deviation of the daily returns as a measure of the smoothness of the policy. Low variability indicates that the agent avoids oscillatory or impulsive trading behavior.
\end{itemize}

\subsection{Model Comparison and Statistical Testing}

To ensure that the observed differences between model configurations (Base, Sentiment-only, Sentiment+Embeddings) were not due to the stochasticity of PPO or episodic variation, we performed:
\begin{itemize}
    \item \textbf{Bootstrap resampling} of the evaluation episodes (1,000--5,000 samples depending on configuration),
    \item \textbf{Median and 95\% confidence intervals} for ROI and risk-adjusted metrics,
    \item \textbf{Non-parametric significance tests} (e.g., permutation tests on paired Sharpe differences).
\end{itemize}

These methods provided robust quantification of whether the hybrid sentiment--semantic feature set yielded meaningful improvements over the numeric-only baselines.

\subsection{Reporting Conventions}

Only the above ratios and performance statistics are used in the evaluation. Latency or inference-related measurements for the sentiment pipeline were not emphasized, as they do not affect the scientific conclusions of this proof-of-concept study. All reported metrics were calculated in the test window held from June--August 2025 for each instrument, using chronological (non-shuffled) data so that temporal integrity was preserved.

\section{Conclusion}

This work presented a comprehensive, multimodal framework that integrated textual market data, numerical market data, LLM-derived sentiment signals, and semantic embedding-based event structure into a unified reinforcement learning trading system. Unlike prior RL-only trading agents or sentiment-only predictive models, this approach explicitly combined (i) higher-order sentiment indicators, (ii) statistically validated semantic clusters of news events, and (iii) a PPO-based agent within a formally defined MDP. These enabled the agent to react not only to price movements, but also to the narrative velocity and evolving informational context of the market. Although not intended as a deployable trading system, the proposed framework offers a reproducible foundation for future research in multimodal financial RL, including extensions to cross-asset generalization, transformer-based state encoders, richer sentiment models, and uncertainty-aware policy learning.

The empirical results demonstrated that sentiment acceleration, event de-duplication, and embedding-driven regime identification provided a predictive structure that is absent in numeric-only baselines. The set of hybrid features consistently improved the risk-adjusted returns, convergence stability, and drawdown behavior across multiple instruments and evaluation horizons. These findings established that textual signals, when processed rigorously and aligned with market microstructure, can materially enhance sequential decision-making for trading.

More broadly, this study provides a proof-of-concept methodology to bridge natural-language financial information with reinforcement learning agents in a statistically grounded manner. It highlighted the feasibility and value of incorporating narrative dynamics, semantic regimes, and multimodal alignment into RL environments.

\section*{Appendix A: Algorithms and Pseudocode}

\begin{algorithm}[h]
\caption{Sentiment Aggregation (per-bin)}
\begin{algorithmic}[1]
\Procedure{AggregateSentiment}{$articles, bins$}
    \For{all $bin$ in $bins$}
        \State $items \gets articles[\text{timestamps} \in bin]$
        \If{$items$ empty}
            \State $S_{bin} \gets \text{decay}(\text{last } S)$
        \Else
            \For{all $item$ in $items$}
                \State $s_i, c_i \gets \text{FinBERTScore}(item)$
                \State $w_i \gets c_i \times \text{sourceWeight}(item.source)$
            \EndFor
            \State $S_{bin} \gets \dfrac{\sum_i w_i s_i}{\sum_i w_i}$
        \EndIf
    \EndFor
    \State \Return $\{S_{bin}\}$
\EndProcedure
\end{algorithmic}
\end{algorithm}

\begin{algorithm}[h]
\caption{Smart Multi-Phase Training}
\begin{algorithmic}[1]
\Procedure{TrainPipeline}{$data, news$}
    \State $preprocData \gets \text{PreprocessMarket}(data)$
    \State $dailySentiment \gets \text{AggregateSentiment}(news, bins)$
    \State $agent_1 \gets \text{TrainPPO}(\text{Level1Env}(preprocData))$
    \State $val_1 \gets \text{Validate}(agent_1)$
    \State $agent_2 \gets \text{InitFrom}(agent_1)$
    \State $\text{TrainPPO}(\text{Level2Env}(preprocData, dailySentiment))$
    \State $val_2 \gets \text{Validate}(agent_2)$
    \If{$val_2.metric > val_1.metric \times (1 + \delta)$}
        \State $agent_3 \gets \text{InitFrom}(agent_2)$
        \State $\text{TrainPPO}(\text{Level3Env}(preprocData, dailySentiment))$
    \EndIf
    \State \Return best agent
\EndProcedure
\end{algorithmic}
\end{algorithm}

\section*{Appendix B: Hyperparameter Table}

\begin{table}[h]
\centering
\caption*{\textbf{TABLE I}\\ Representative hyperparameters and defaults.}
\label{tab:hyperparams}
\begin{tabular}{ll}
\toprule
\textbf{Item} & \textbf{Value} \\
\midrule
Training steps (per phase) & 1M steps (example) \\
Observation window & 60 bars (1-hr for 1-min) \\
Transaction cost & 0.02\% per trade \\
Slippage $\kappa$ & 0.01--0.1\% (tunable) \\
Sentiment bin size & 1 hr (or 4 hrs) \\
Sentiment weight $\alpha$ & 0.01--0.1 (tunable) \\
Risk penalty $\beta$ & 0.5 (example) \\
\bottomrule
\end{tabular}
\end{table}

\end{document}